\documentclass[sigconf,nonacm]{acmart}
\AtBeginDocument{%
  \providecommand\BibTeX{{%
    \normalfont B\kern-0.5em{\scshape i\kern-0.25em b}\kern-0.8em\TeX}}}
\usepackage{caption}
\usepackage{epsfig}
\usepackage{setspace}
\usepackage{balance}
\usepackage{algorithm}
\usepackage{algorithmic}
\usepackage{framed}
\usepackage{color}
\usepackage{xcolor}

\usepackage{subcaption}

\newcommand{\eat}[1]{}
\newcommand{\zhaqieat}[1]{}

\def\P{\mathbb{P}}
\def\R{\mathbb{R}}

\def\1{\mathbf{1}}

\newcommand{\argmin}[1]{{\underset{#1}{\operatorname{argmin}}}}

\begin{document}
\title{Active Learning for Skewed Data Sets}

\author{Abbas Kazerouni}
\affiliation{%
  \institution{Stanford University}
  \streetaddress{450 Serra Mall}
  \city{Stanford}
  \country{US}
}
\email{abbas.kazerouni@gmail.com}

\author{Qi Zhao} 
\affiliation{ 
  \institution{Google}
  \streetaddress{1600 Amphitheatre Parkway}
  \city{Mountain View} 
  \country{US}
}
\email{zhaqi@google.com}

\author{Jing Xie} 
\affiliation{ 
  \institution{Google}
  \streetaddress{1600 Amphitheatre Parkway}
  \city{Mountain View} 
  \country{US}
}
\email{lucyxie@google.com}

\author{Sandeep Tata} 
\affiliation{ 
  \institution{Google}
  \streetaddress{1600 Amphitheatre Parkway}
  \city{Mountain View} 
  \country{US}
}
\email{tata@google.com}

\author{Marc Najork} 
\affiliation{ 
  \institution{Google}
  \streetaddress{1600 Amphitheatre Parkway}
  \city{Mountain View} 
  \country{US}
}
\email{najork@google.com}


\begin{abstract}
Consider a sequential active learning problem where, at each round, an agent selects a batch of unlabeled data points, queries their labels and updates a binary classifier.
While there exists a rich body of work on active learning in this general form, in this paper, we focus on problems with two distinguishing characteristics: severe class imbalance (skew) and small amounts of initial training data.
Both of these problems occur with surprising frequency in many web applications. 
For instance, detecting offensive or sensitive content in online communities (pornography, violence, and hate-speech) is receiving enormous attention from industry as well as research communities.
Such problems have both the characteristics we describe -- a vast majority of content is {\em not} offensive, so the number of positive examples for such content is orders of magnitude smaller than the negative examples.
Furthermore, there is usually only a small amount of initial training data available when building machine-learned
models to solve such problems.
To address both these issues, we propose a hybrid active learning algorithm (HAL) that balances exploiting the knowledge available
through the currently labeled training examples with exploring the large amount of unlabeled data available.
Through simulation results, we show that HAL makes significantly better choices for what points to label
when compared to strong baselines like margin-sampling.
Classifiers trained on the examples selected for labeling by HAL easily out-perform the baselines
on target metrics (like area under the precision-recall curve)
given the same budget for labeling examples.
We believe HAL offers a simple, intuitive, and computationally tractable way to structure
active learning for a wide range of machine learning applications.
\end{abstract}
\maketitle

\section{Introduction}
A key step in learning high-quality models in a variety of supervised learning scenarios is obtaining labeled training examples.
Several applications start with a large number of unlabeled examples, and need to acquire
labels for training by presenting the examples to humans for judgment.
This can be expensive since it might require setting up tools and infrastructure, training
humans on the evaluation task, and paying people for the time spent labeling the examples.
Algorithms that select examples for labeling
that are likely to give us the most improvement are clearly valuable.
With a good algorithm, we will be able to obtain the same amount
of improvement to a target metric for lower cost, or obtain greater improvements
at the same cost compared to a na\"{\i}ve algorithm.

In this paper, we consider the problem of designing a good active learning algorithm~\cite{settles2012active}
for learning a binary classifier on a highly skewed data set while starting out with a
limited amount of training data.
This is a surprisingly common scenario for many applications.
Consider building a spam model on an open discussion platform on the web.
The data is often highly skewed, with most examples being non-spam. 
For applications where we want to detect sensitive or offensive content (pornography,
hate-speech) we face a similar problem -- most data is in the
negative class (non-offensive), with
a very small number of examples in the positive class. 
\eat{In this paper, we focus on binary classifiers for simplicity, but the ideas are
more generally applicable.}
Furthermore, we focus on the difficult setting where we are starting out building the
classifier and have very little training data (in the limit, we have no labeled
examples at all).

A well-known active learning baseline is margin sampling~\cite{balcan2007margin}.
The key intuition in margin sampling is to sample from unlabeled points with a
probability inversely proportional to the margin (distance from the separating
hypersurface). Empirical studies~\cite{schein2007active} have shown this approach to be surprisingly effective
in a variety of settings.
In this paper, we present a novel active learning algorithm, called Hybrid Active Learning (HAL), that leverages an explore-exploit
trade-off to improve on margin sampling.
The key insight is to combine margin sampling, a strategy that exploits the existing
labeled data for incremental improvements to the training data set, with an exploration
scheme that allows us to improve the classifier quicker. Margin sampling selects new points based on previously labeled points which
could potentially introduce bias to the training set.
Furthermore, margin sampling may get stuck at a particular uncertain area of the whole input space and leave out other unexplored areas.
By sampling from unexplored areas in the input space, the exploration scheme exposes new uncertain areas to the margin sampler and improves its usefulness. 

Our proposed algorithm allows for a generic exploration scheme to be combined with margin sampling. Hence, the computational complexity of our algorithm depends on the complexity of the exploration scheme. As will be presented later in this paper, very simple exploration schemes such as random and Gaussian exploration significantly improve over margin sampling.

The paper makes the following key contributions:
\begin{itemize}
    \item It presents HAL, a novel active learning scheme that leverages the explore-exploit trade-off for active learning on highly skewed data sets.
    \item It demonstrates through studies on real and synthetic data that the proposed algorithm significantly outperforms margin sampling for our experimental setting with skewed data.
    The advantage is particularly large during the initial stages where we have much less training data.
    \item It identifies interesting avenues for further research into identifying better active learning schemes over highly skewed data sets.
\end{itemize}

The rest of the paper is structured as follows. Section~\ref{sec::relatedwork} surveys areas of research
in active learning. Section~\ref{sec::problemFormulation} formally introduces the problem setting
and describes the key challenges of skew and limited training data.
Section~\ref{sec::proposed_algorithm} describes the intuition behind our approach and the details
of the exploration schemes we consider. Section~\ref{sec::results} presents experimental results
on synthetic and real datasets and compares our approach to multiple baselines. Section~\ref{sec::discussion} discusses various aspects of the proposed algorithms and identifies future research directions for designing better active learning algorithms. Finally, we summarize our findings and conclude the paper in Section~\ref{sec::conclusions}.

\section{Related Work}
\label{sec::relatedwork}
There exists a vast body of work on active learning in various scenarios. We refer to \cite{fu2013survey} and \cite{settles2012active} for an extensive review of the literature. In this paper, we are considering a pool-based active learning problem where an agent has access to a large set of unlabeled examples whose labels can be obtained through human annotation. In such a setting, active learning algorithms proposed in the literature select points based on two principles. Most algorithms are based on the informativeness principle which selects the most informative points. Two popular approaches are margin sampling \cite{lewis1994sequential, lewis1994heterogeneous, balcan2007margin} and query-by-committee \cite{seung1992query, freund1997selective, dagan1995committee}.
The former algorithm selects unlabeled points for which the current classifier is most uncertain. The latter picks points about which multiple trained classifiers 
most disagree. One key feature of these types of algorithms is that the decision about what example should be labeled next is solely based on the previously labeled examples.
This could potentially lead to a high bias in the training set, specially if only a few labeled examples are initially available. The second principle in active learning is to select examples that are most representative of the whole unlabeled data set. This principle forces the algorithms to exploit the spatial structure of the unlabeled data set, for instance through clustering \cite{nguyen2004active, dasgupta2008hierarchical}.  As such, the performance of such algorithms is heavily influenced by the quality of the clustering algorithm. Many research approaches combine the two above principles. \cite{xu2003representative} and \cite{donmez2007dual} mix between the two principles and \cite{huang2010active} propose an algorithm based on the min-max view of active learning.

Despite the existence of  many active learning algorithms in the literature, none of them is believed to be superior to the others in a general scenario. \cite{ramirez2017active} provides extensive simulation results comparing different active learning baselines, used with various classifiers, across multiple data sets. One key observation made by this empirical study is that the best active learning algorithm varies by the data set and even by the performance measure of interest. 

When dealing with highly imbalanced classes, \cite{attenberg2010label} has proposed guided learning where instead of only querying the label for an unlabeled data point, the agent asks crowd workers to find or generate a training example from the minority class. Although such a strategy ensures the existence of enough examples of each class in the training set, it requires expensive human effort. To alleviate the high cost of guided learning, \cite{c2018active} proposes to mix between example generation and label querying. Our algorithm is an alternative low-cost solution to the skewed data problem which does not require the costly example generation task. In \cite{ertekin2007learning}, the authors propose to select samples within within the SVM margins and they argue that such samples are less imbalanced compared to the entire dataset. As sampling within margins is similar to margin-sampler algorithm, their approach is likely to suffer from the same issue illustrated in Figure\ref{fig::margin} as margin-sampler algorithm.

In \cite{brinker2003incorporating, xu2007incorporating}, the authors propose to consider the diversity of the samples selected for labeling. As their approaches are motivated for general active learning problems, it's unclear how they perform in problems where the class is extremely imbalanced.

Another approach to exploit unlabeled data when training a classifier is via semi-supervised learning techniques \cite{chapelle2009semi, zhu2011semi}. It is worth mentioning that these approaches are orthogonal to our aim in this paper. In particular, our goal is to design active learning algorithms that select which unlabeled points should be labeled next. Still, when updating the classifier at each round, semi-supervised techniques can be leveraged to come up with a better classifier. Further investigating the combination of active and semi-supervised learning is an interesting area of future research.

The crowd-computing community has a rich literature on studying
effective and efficient ways to leverage human effort
for computational tasks.
For example, marketplaces like Amazon's Mechanical Turk (MTurk)
make it easy to distribute such tasks that require human
intelligence and make it simple, scalable, and cost-effective. 
Several research efforts have leveraged such services in image classification~\cite{sorokin2008utility, deng2009imagenet, su2012crowdsourcing}, machine translation~\cite{ambati2010active}, etc. The methods in this literature can be divided into two general categories. One category is the conventional active learning framework where an algorithm selects unlabeled data points and human intelligence is employed to label this selected point \cite{donmez2009efficiently, wallace2011should, kajino2015active, yang2018leveraging, moon2014active}. The other category consists of methods that rely more on human intelligence. In particular,  human resource may be employed to generate a certain class of data points, or annotate data points beyond only providing a label. For instance, \cite{margineantu2005active, haertel2008return, arora2009estimating, hu2018active} have considered the possibility of different types of queries with different informativeness and cost and in  \cite{moon2014active}, the researchers use guided learning techniques to generate new data points aside from labeling the existing unlabeled ones. While methods in the second category allow for more informative actions, they require a balanced trade-off between the informativeness and cost of a query. Furthermore, such methods are vulnerable to other issues as human generated examples may introduce bias into the training data set. The algorithms in this paper belong to the first category which employs minimal human intelligence.  

Several ideas to improve data quality by using multiple labelers~\cite{sheng2008get, welinder2010online, welinder2010multidimensional, snow2008cheap}
and estimating labeler expertise and reliability are clearly relevant and can be used to further improve the algorithms proposed in this paper.

In this paper, we propose a hybrid active learning algorithm which, at each round, trades off between exploiting the currently learned classifier and exploring undiscovered parts of the input space. The exploration-exploitation dilemma has been widely studied in bandit and reinforcement learning literature \cite{gittins2011multi, sutton2018reinforcement, thrun1995exploration} and many algorithms have been designed for different settings. The idea of combining exploration and exploitation applies to active learning as well. \cite{osugi2005balancing} studies balancing between exploration and exploitation for active learning algorithm by dynamically adjusting the probability to explore at each step. \cite{cebron2009active} proposes a prototype based active learning algorithm by leveraging exploration/exploitation strategy based on uncertainty distribution. But both of them only focus on general active learning problem. As mentioned in \cite{attenberg2011inactive}, main challenges when applying active learning in practice including dealing with the situation when data distribution is highly skewed and with highly disjunctive classes. Our approach is going to tackle these difficulties.

\begin{figure}
\centering
\begin{subfigure}{.23\textwidth}
  \centering
  \includegraphics[width=.95\linewidth]{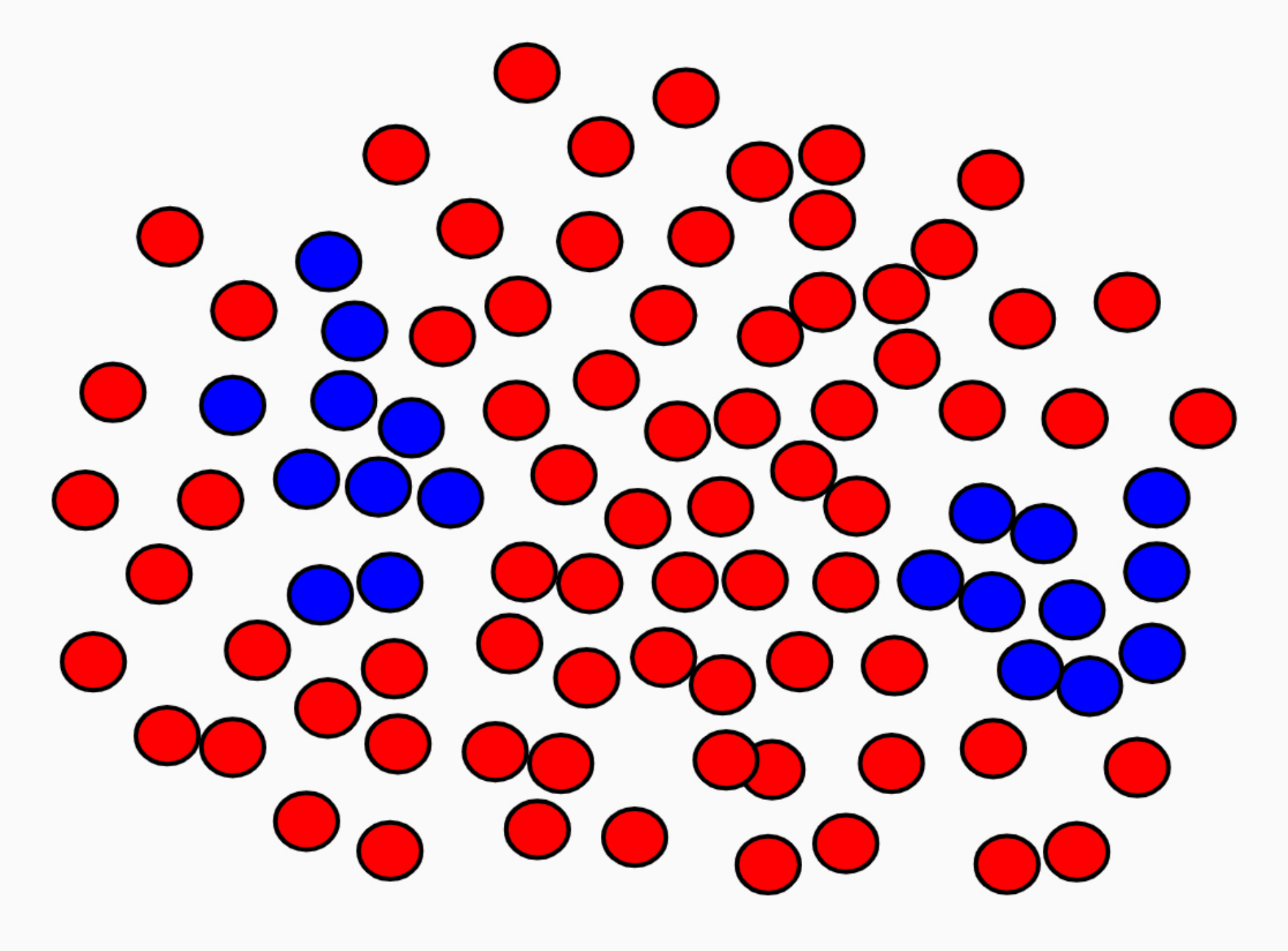}
  \caption{All examples with their labels}
  \label{fig::all_data}
\end{subfigure}%
\begin{subfigure}{.23\textwidth}
  \centering
  \includegraphics[width=.95\linewidth]{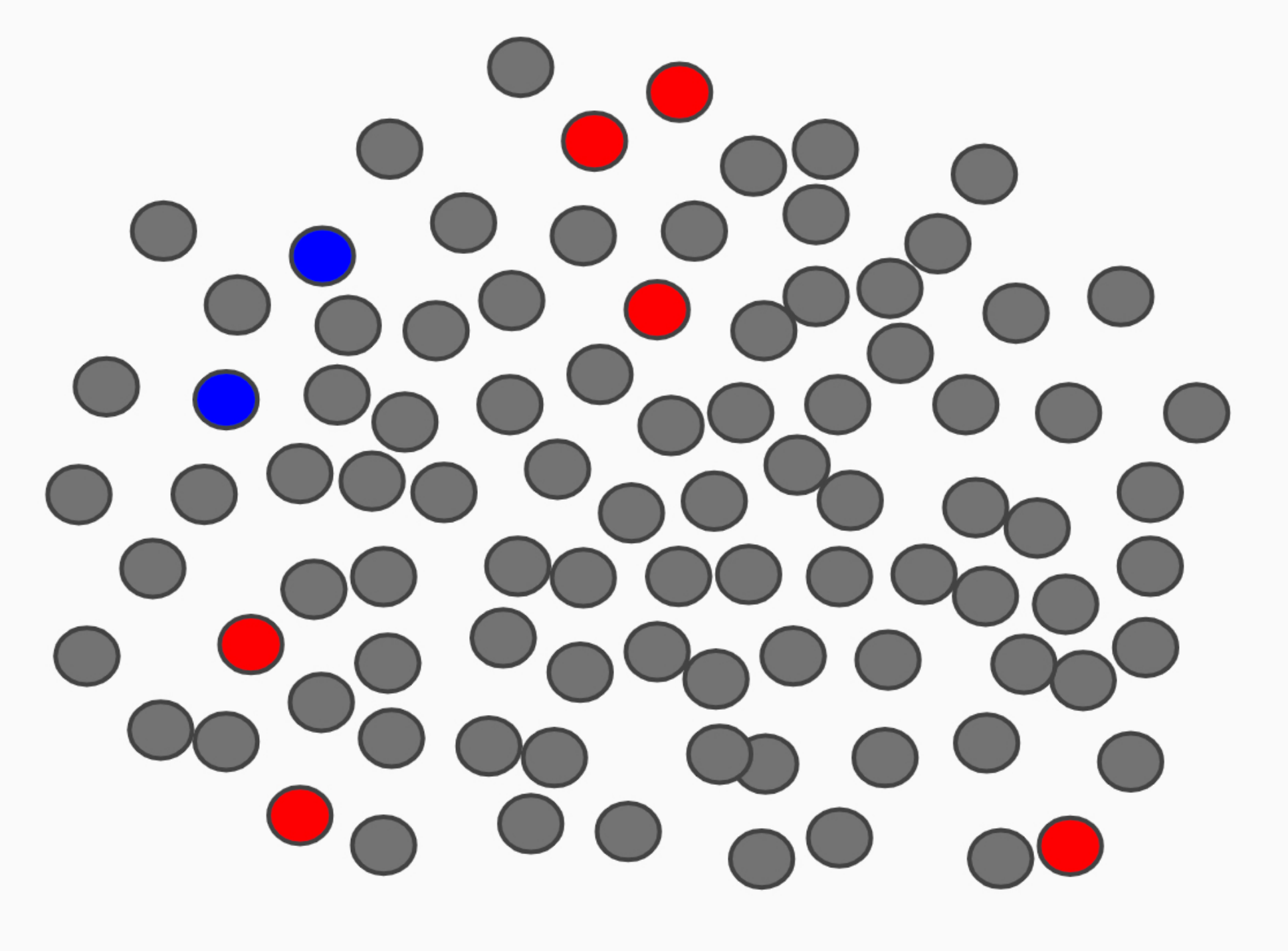}
  \caption{Starting labeled data set}
  \label{fig::starting_set}
\end{subfigure}
\caption{(a) The input space with their true label and (b) the initial labeled data set $L_0$. Blue and red points represent two different labels, say positive and negative, respectively, while gray points represent unlabeled examples.}
\label{fig::all_and_starting}
\vspace{-3mm} 
\end{figure}

\section{Problem Formulation}
\label{sec::problemFormulation}
Consider a classification problem where a feature vector $x\in\R^d$ is mapped to a label $y$ in a finite set of possible labels  $\{1,2,\cdots,K\}$. An agent has access to a large set of unlabeled data along with a smaller set of labeled data based on which a classifier can be trained. The agent's goal is to maximally improve the classifier's performance on a target metric using a given labeling budget. To that end, the agent repeatedly selects a subset of the unlabeled data set, obtains the corresponding labels, and retrains the classifier using the expanded labeled data. More formally, let $U_t = \{x_i\}_{i=1}^{u_t}$ denote the unlabeled data set at time step $t$ and let $L_t = \{(x_i,y_i)\}_{i=1}^{l_t}$ denote the labeled data set at that time. Based on these two data sets, the agent may employ supervised or semi-supervised learning methods to train a classifier $C_t$ at time $t$. At each time step $t$, the agent is allowed to select a set $M_t$ consisting of $m$ unlabeled data points in $U_t$ and query their labels to get a set of $m$ labeled points $\bar M_t$. Then, the data sets at the next time step are $U_{t+1} = U_t - M_t$ and $L_{t+1} = L_t\cup \bar M_t$ and a new classifier $C_{t+1}$ is trained on $L_{t+1}$. This process repeats for a sequence of time steps $t=0,1,2,\cdots,T$. 

Generally, the extra information acquired by labeling the points $M_t$ will improve the classifier's performance. However, the agent's goal is to intelligently select the set of points to be labeled such that the classifier's performance improves most at a given labeling cost. Depending on the application, the performance of the classifier can be measured in terms of different metrics such as its accuracy, area under precision-recall curve and recall at a certain precision. In this paper, we assume that there is no access to a validation set when the active learning algorithm is being deployed, and hence, the algorithm cannot depend on the feedback it receives by evaluating the classifier on the validation set. 

We make two additional assumptions that distinguish our problem from the standard active learning problem. 
First, we assume that the data set is skewed; that is, the majority of data points belong to one of the possible classes.
This is a common scenario in many real-world applications like detecting sensitive content (porn, hate-speech) or spam and phishing attacks in online communities.
With this constraint, additional effort is needed to ensure that the active learning algorithm adds sufficiently many data points from the minority classes to the training set. 
Second, we assume that the agent starts with very few labeled data points (i.e., very small $L_0$). 
This is common in every real-world setting of building a new model from scratch.
The classifier is extremely unreliable during the initial phases and the active learning algorithm cannot rely solely on the classifier's predictions.
These two problems call for a more sophisticated active learning algorithm.

\section{Proposed Algorithm}
\label{sec::proposed_algorithm}
To address the challenges discussed in Section \ref{sec::problemFormulation}, we design an active learning algorithm consisting of two main components, called {\em exploit} and {\em explore}. By mixing between these two components, the algorithm outperforms margin sampling on data with highly imbalanced classes and with a very small initial training set.

\subsection{High-Level Intuition}
\label{subsec::intuition}
Before presenting the details, let us describe the high level intuition through a visual example. Consider a binary classification problem in a skewed environment where all examples and their true labels are as depicted in Figure \ref{fig::all_data}. Suppose that the agent starts with only 8 labeled examples as depicted in Figure \ref{fig::starting_set} and assume that it is allowed to query labels for 4 points at each round. 

\begin{figure}
\centering
\begin{subfigure}{.33\columnwidth}
  \centering
  \includegraphics[width=.95\linewidth]{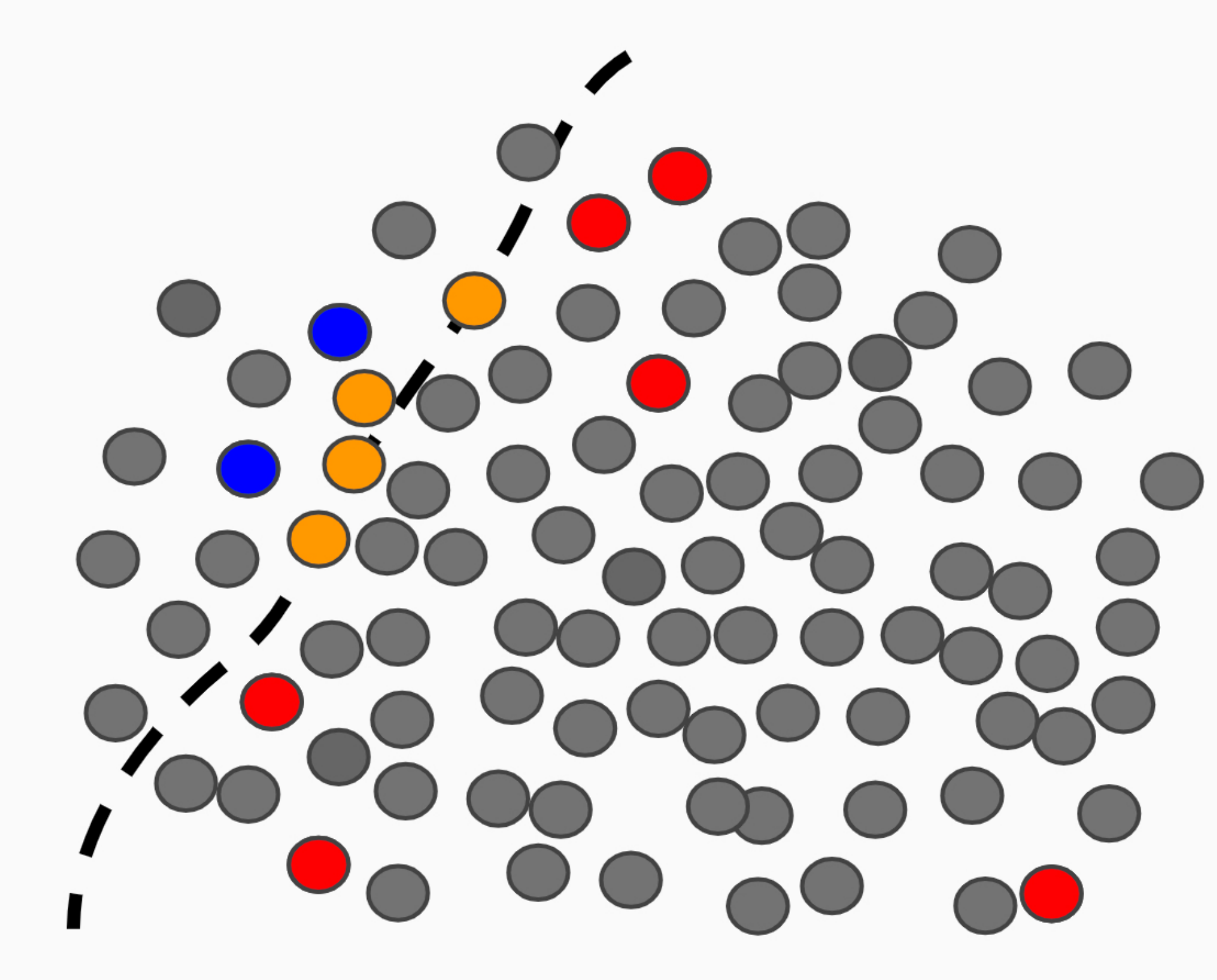}
  \caption{}
  \label{fig::margin1}
\end{subfigure}%
\begin{subfigure}{.33\columnwidth}
  \centering
  \includegraphics[width=.95\linewidth]{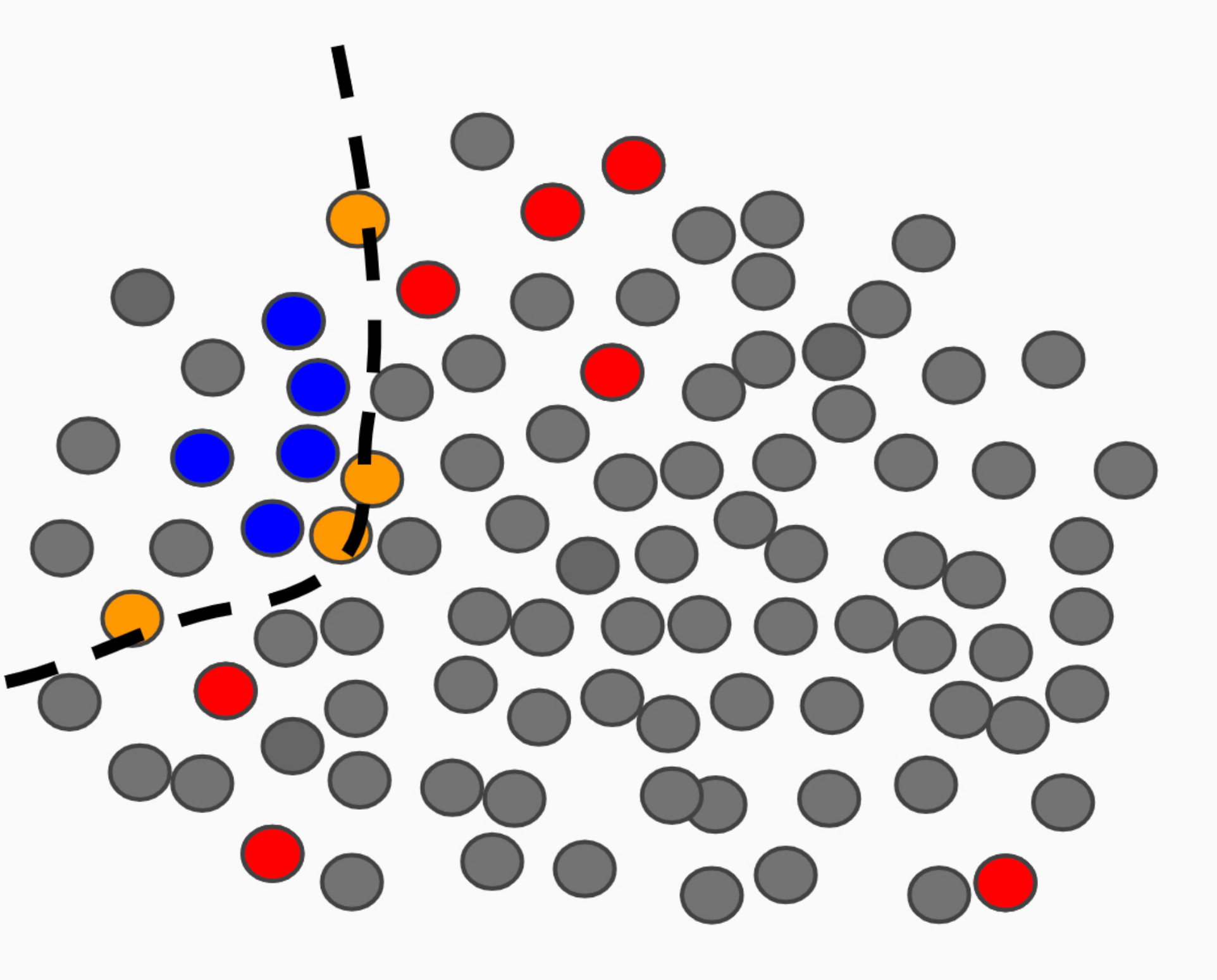}
  \caption{}
  \label{fig::margin2}
\end{subfigure}
\begin{subfigure}{.33\columnwidth}
  \centering
  \includegraphics[width=.95\linewidth]{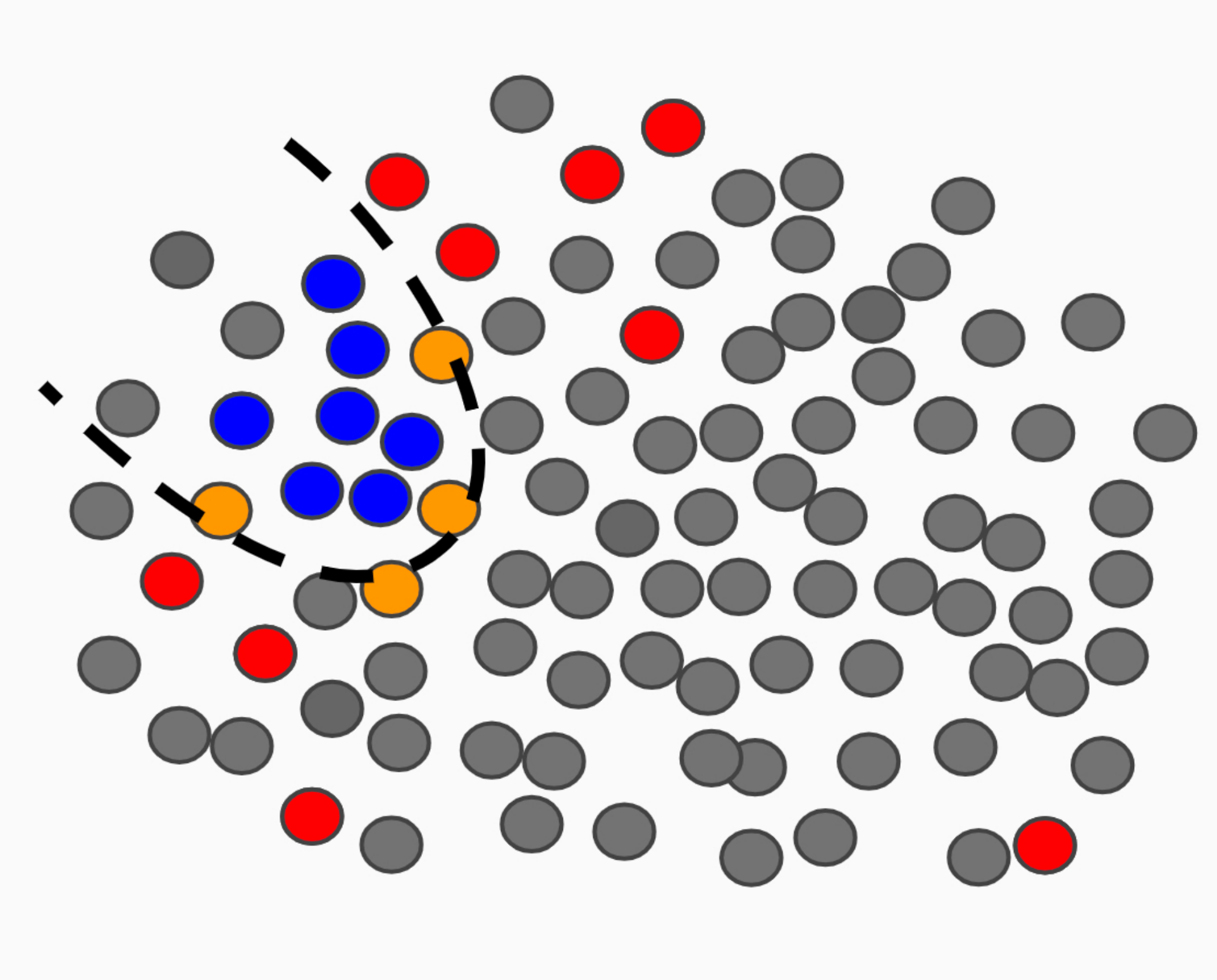}
  \caption{}
  \label{fig::margin3}
\end{subfigure}
\caption{Examples selected by margin sampler at successive rounds. Orange points represent selected data points. The dashed line represents a hypothetical separating hypersurface.}
\label{fig::margin}
\vspace{-6mm} 
\end{figure}

Figure \ref{fig::margin} shows examples obtained by margin sampling at three successive rounds. Recall that margin sampling picks the most uncertain points at each round; i.e., the points closest to the separating hypersurface. As such, margin sampling starts by selecting 4 points as in Figure \ref{fig::margin1}. After obtaining labels for these 4 points, the separating hypersurface is updated as in Figure \ref{fig::margin2} and the next 4 points are selected accordingly. As can be seen in Figure \ref{fig::margin3}, margin sampling keeps selecting points in the same uncertain area, and it may take a very long time to discover the other positive cluster of points on the right side of the input space (Figure \ref{fig::all_data}). As a result, the margin sampler picks a poor collection
of points for the training set.

\begin{figure}
\centering
\begin{subfigure}{.33\columnwidth}
  \centering
  \includegraphics[width=.95\linewidth]{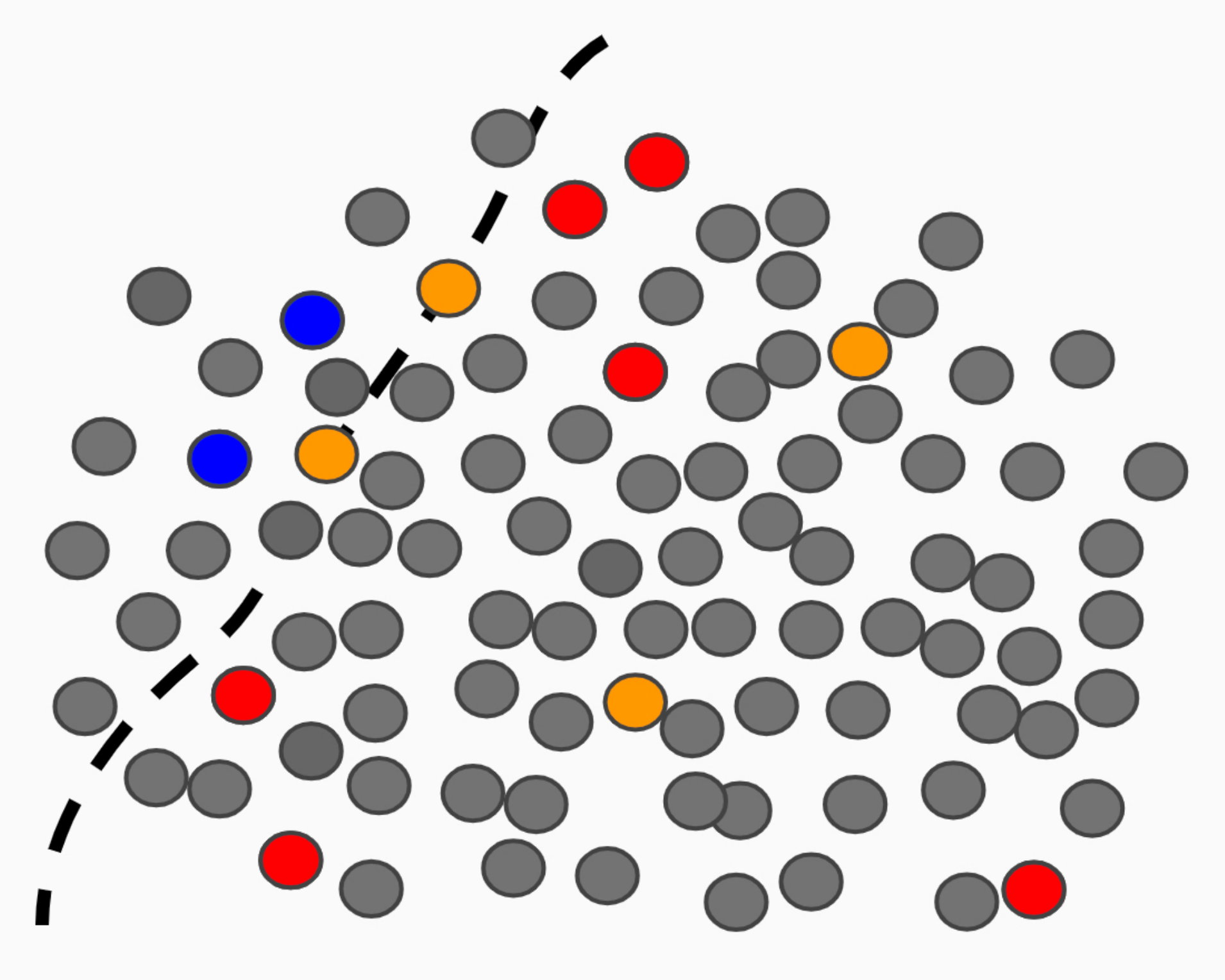}
  \caption{}
  \label{fig::hybrid1}
\end{subfigure}%
\begin{subfigure}{.33\columnwidth}
  \centering
  \includegraphics[width=.95\linewidth]{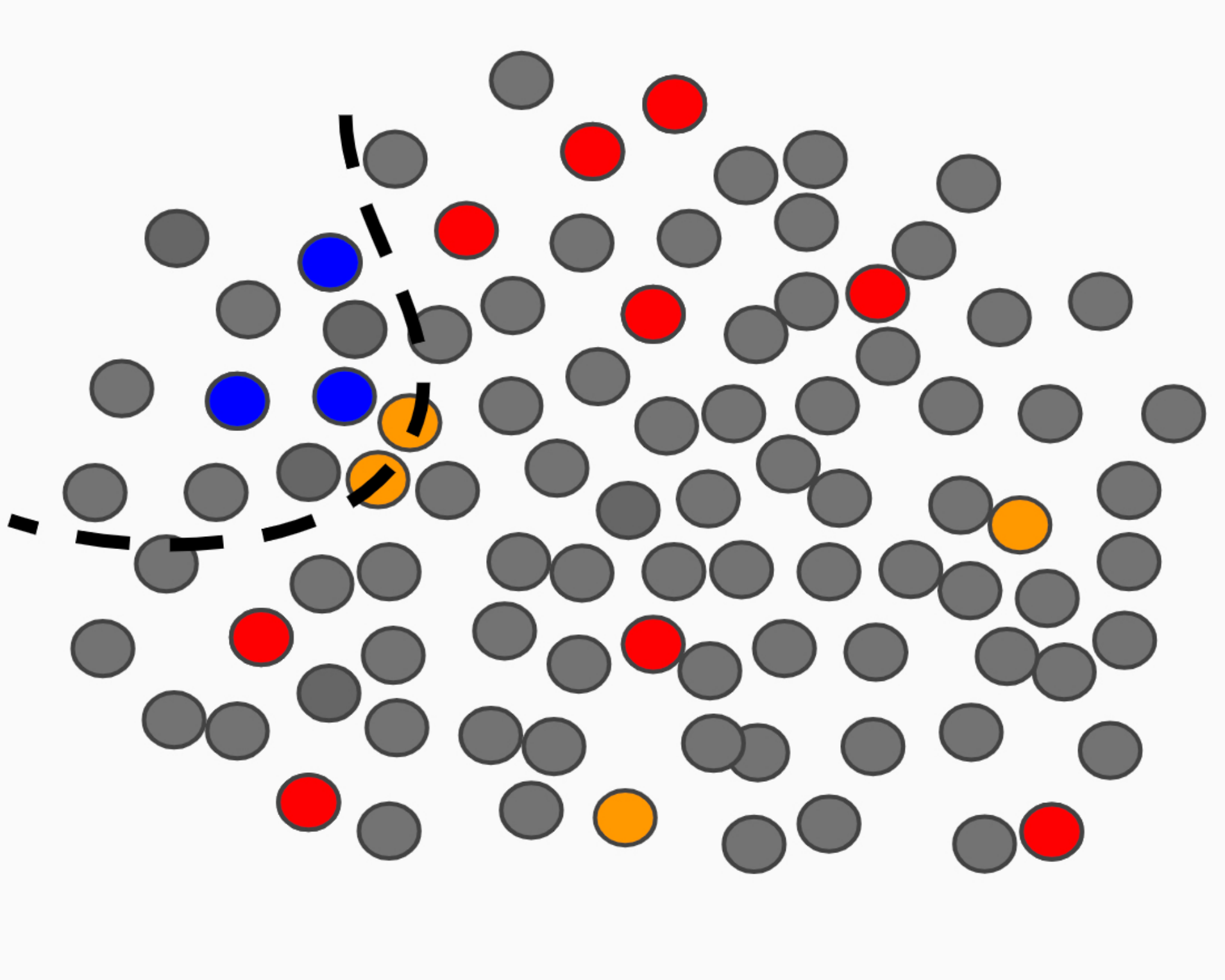}
  \caption{}
  \label{fig::hybrid2}
\end{subfigure}
\begin{subfigure}{.33\columnwidth}
  \centering
  \includegraphics[width=.95\linewidth]{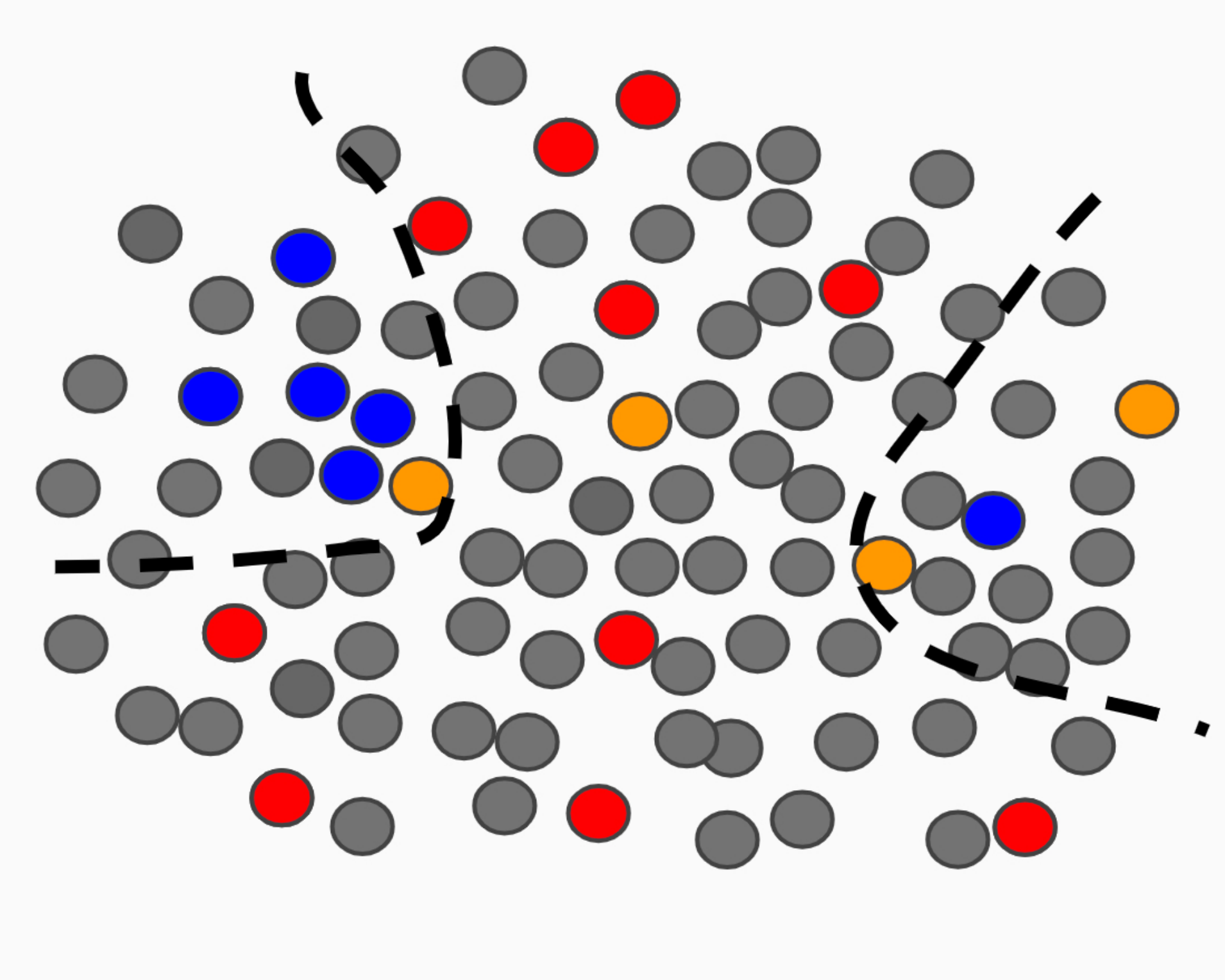}
  \caption{}
  \label{fig::hybrid3}
\end{subfigure}
\caption{Successive samples selected by a hybrid algorithm which selected 2 points according to margin sampler and 2 points according to an exploration scheme which picks points from the less explored areas of the input space.}
\label{fig::hybrid}
\vspace{-3mm}
\end{figure}

Figure \ref{fig::hybrid} shows successive samples taken by a hybrid algorithm which selects 2 examples based on margin sampling and 2 points according to an exploration scheme which selects points from the least explored areas. As depicted here, such a hybrid algorithm discovers the other positive cluster of points very quickly. Once one of these positive points is exposed, a new uncertain area forms around it and hence at later rounds, the margin sampler will discover even more positive points in that area.

The above example illustrates the key shortcoming of margin sampling and other exploit-only active learning algorithms. Furthermore, it shows how combining an exploration scheme with margin sampling can significantly improve the performance. Given this high-level intuition behind our algorithm, let us describe each component in more details.

\subsection{Exploit Component}
\label{sec::exploit_component}
This component of the algorithm is based on the predictions of the current classifier. In other words, at time $t$, this component exploits  $C_t$'s predictions on each of the unlabeled points when deciding about the set $M_t$. While there might be different ways to exploit the classifier's prediction, we take it to be margin sampling. At each time $t$, margin sampling selects points the label of which the classifier $C_t$ is most uncertain about. Specifically, classifier $C_t$ suggests a prediction vector $\pi_t(x)=(\pi_t^1(x), \pi_t^2(x),\cdots,\pi_t^K(x))$ for each unlabeled point $x\in U_t$ such that $\pi_t^k(x)$  denotes the probability of $x$ being of class $k$. Given this, we can define a certainty score for each point $x\in U_t$ as
\begin{equation}
\label{certaintyScore}
c_t(x) = |\pi_t^{(1)}(x) - \pi_t^{(2)}(x)|,
\end{equation}
where $\pi_t^{(1)}(x)$ and $\pi_t^{(2)}(x)$ are the maximum and second maximum components of $\pi_t(x)$, respectively.
The certainty score represents how certain the classifier $C_t$ is about $x$'s label. When the classifier is certain about a point $x$ to be of one of the possible classes, the certainty score in \eqref{certaintyScore} is large. On the other hand, if the classifier is not confident about what label $x$ should have the prediction probabilities are close together resulting in a small certainty score. Given the certainty scores for each unlabeled point, the margin sampler selects the points with lowest certainty score. 

\subsection{Explore Component}
As mentioned in Section \ref{sec::problemFormulation}, the classifier is likely to be unreliable during the first phases when it has not yet been fed with enough data points. Therefore, solely relying on the exploit component may result in poor performance of the algorithm. In particular, the margin sampler picks the points close to the decision boundary and hence focuses on a limited area of the whole space. 

To address this issue, the explore component is designed to select points in the unexplored areas, independent of the classifier's prediction. Specifically, at each time $t$, an exploration score $s_t(x)$ is assigned to each unlabeled point $x\in U_t$ that measures how explored the area around $x$ is. Then, the exploration component picks the points with the smallest exploration score. There are many ways to assign exploration scores to unlabeled points. In this paper, we discuss three such schemes.  

\textit{Random exploration} simply selects the unlabeled points uniformly at random. This is equivalent to assigning random exploration score $s_t(x)$ to each point $x \in L_t$ at round $t$, or more formally,
\begin{equation}
s_t^R(x)\sim \mathit{uniform}(0,1)
\end{equation}

\textit{Gaussian exploration}, being more sophisticated, works by assigning the following score to each unlabeled point $x$ at round $t$:
\begin{equation}
\label{softExpScore}
s^{G}_t(x) = \sum_{z\in L_t}\exp\left(-\frac{\|x-z\|_2}{\delta}\right).
\end{equation}
Here, $\delta$ serves as a free parameter of the Gaussian exploration scheme which governs the effect of a labeled point on the exploration score of the unlabeled points. Clearly, the closer an unlabeled point $x$ to a labeled point $z$, the larger the score. The score is a reasonable measurement of the certainty of the label for $x$ if we assume that the label (or function value in a general sense) changes smoothly in the feature space. This can be better understood by considering a simple case with two close-by points $a$ and $b$ where the label for $a$ is known. Since $b$ is close to $a$, it tends to have the same label as $a$ due to space smoothness assumption. In this case, knowing $b$'s label adds little information to the training data. As a result, we should select points with lower exploration scores for labeling. 

A third possible exploration scheme is to consider a {\em neighborhood} around each unlabeled point and take the exploration score to be the fraction of labeled points in that area. For example, the neighborhood for each point could be the set of the $N$ closest points to it. According to our experiments, such an exploration scheme does not perform as well as the two previous ones, and we do not report it in the simulation results section. It is still possible to define more sophisticated, perhaps dynamically changing, neighborhoods that give rise to more representative exploration scores. We consider this as a possible topic for future research.

One important consideration when designing the exploration scheme is its computational complexity. Specifically, since scores are being computed for each unlabeled points at every round, there should be a computationally easy way to update them. Otherwise, our algorithm would be prohibitively expensive. We discuss this in more detail in Section \ref{sec::computational}.

\subsection{Final Algorithm}
Now that we have discussed the two components of our algorithm, we can present the full algorithm combining their advantages.

Our algorithm, described in Algorithm \ref{alg::finalAlg}, has a parameter $p\in [0,1]$ which denotes the trade-off between the explore and exploit components. At each round $t$, the algorithm has a budget to select $m$ points. Each point is picked according to the exploit or explore component with probability $p$ and $1-p$, respectively. Once a point is selected, the exploration score is updated for the remaining unlabeled points and this process repeats until all $m$ points have been selected.

We name our algorithm Hybrid Active Learning (HAL). In the rest of the paper, we use HAL-R($p$) and HAL-G($p$) to denote HAL with random and Gaussian exploration and with a trade-off parameter $p$.  Note that HAL-R(1) is equivalent to margin sampling and HAL-R(0) is equivalent to random sampling.

\begin{algorithm}[tb]
   \caption{Hybrid Active Learning (HAL)}
   \label{alg::finalAlg}
\begin{algorithmic}
   \STATE {\bfseries Input:} Initial sets $U_0, L_0$; labeling budget $m$; trade-off parameter $p$;  exploration scheme $s$, exploitation scheme $c$
   \STATE{\bfseries Initialize:} $U=U_0,L=L_0$ and compute $s(x)$ for $x\in U_0$
   \FOR{$t=0,1,2,\cdots$}   
   \STATE Update the classifier $C$ based on $U$ and $L$
   \STATE Compute the certainty scores $c(x)~\forall x\in U$
   \STATE Let $M = \{\}$
   \FOR{$i=1,2,\cdots,m$}
       \STATE Let $z = \left\{\begin{array}{ll}
       \argmin{w\in U}~c(w) &\mbox{with probability } p\\
       \argmin{w\in U}~s(w) &\mbox{with probability } 1-p
       \end{array}\right.$ 
       \STATE Remove $z$ from $U$ and add it to $M$
       \STATE Update $s(x)$ for $x\in U$
   \ENDFOR
   \STATE Get the labels for points in $M$
   \STATE Update $L \leftarrow L\cup \bar M$
   \ENDFOR
   \STATE Update the classifier $C$ based on $U$ and $L$
\end{algorithmic}
\end{algorithm} 

\subsection{Computational Complexity}
\label{sec::computational}
We briefly mentioned the computational complexity of the Gaussian exploration algorithm. Now let us give a formal treatment of the complexity of the entire algorithm. Since explore and exploit are separate components, we can derive the complexity for each and simply add them up to get the final complexity. Let $n$ and $m$ denote the total number of data points and the labeling budget at each step, respectively. The complexity at step $t \in [0, \frac{n}{m}]$ and the total complexity of labeling all data points is summarized in Table~\ref{tab:complexity},

\begin{table}
    \caption{Computational complexity of each component at step $t$ and the total complexity for all data points being labeled.}
    \centering
    \begin{tabular}{|l|l|l|}
    \hline
    Component & Step-t & Total \\\hline\hline
    Margin Sampling & $O(n-pmt)$ & $O\left((1-\frac{p}{2})\frac{n^2}{m}\right)$ \\
    Random Exploration & $O(m)$ & $O(n)$ \\
    Gaussian Exploration & $O((1-p)mn)$ &  $O((1-p)n^2)$ \\
    Model Update & $O(mt)$ &  $O(\frac{n^2}{m})$  \\\hline\hline
    \end{tabular}
    \label{tab:complexity}
    \vspace{-3mm}
\end{table}
Note that, on average at each round, margin sampling selects $pm$ points. Using a \textit{Quickselect}\cite{Hoare:1961:AF:366622.366647} based algorithm, it will have a $O(n-pmt)$ complexity
on average.
Updating the classifier at each round results in a $O(mt)$ complexity as the training only goes through the current labeled data a fixed number of times. At each round,  Gaussian exploration picks an average of $(1-p)m$ points sequentially. After selecting each point, the exploration scores of remaining unlabeled points are updated resulting in a $O((1-p)mn)$ complexity at round $t$.
 The complexity for margin sampling combined with any of the two exploration schemes is $O(\frac{n^2}{m})$.

\section{Experiment Results}
\label{sec::results}
In this section, we present simulation results for the case of
binary classification with positive and negative class labels. We compare our
method against baseline algorithms both on synthetic and real-world data sets.
Both these cases represent a highly skewed binary data set where only a small
fraction of data points are positively labeled. Moreover, we consider the
scenario where the active learning process starts with zero labeled data
points. We use a simple neural network for the classifier.
In each case, we use a two hidden layers with $50$ nodes on each layer for the synthetic data set and $20$ nodes on each layer for the MNIST data set. Each layer is fully connected, and uses rectified linear units (ReLU). The network is set up to optimize the cross-entropy loss. Adagrad\cite{duchi2011adaptive} is used to train the network. The initial learning rate is set to $0.05$. We tried different neural network
architectures on the entire labeled data set and selected the optimal
architecture based on the best performance on the testing data set. The active
learning algorithm selects 100 new points from the unlabeled set at each round.
These selected data points are then labeled and added to the training set to
update the classifier.

In each of the following scenarios, we evaluate HAL both for random exploration and Gaussian exploration. For Gaussian exploration, we simply set the scaling factor $\delta$ to $10$. Furthermore, we compare against margin sampling and random sampling as two baseline algorithms.
We report two performance metrics for the classifiers trained
at each time step -- the area under the precision-recall curve (AUC-PR) and the recall at precision of 0.9.
Recall at high precision is a more useful metric in applications where the positive class is very rare, and
the goal in many applications like sensitive-content detection is to recover as many positives as possible at a high precision bar.

\subsection{Synthetic Data Set}\label{sec::synthetic_data}
Let us start by presenting the simulation results on the synthetic data set. The data set consists of 10-dimensional data points which are generated as follows. First, 300 random points are generated in the 10-dimensional space by sampling from a centered multivariate normal distribution with independent components each having a variance of $8$. These points are then considered as  the centers of 300 clusters.  Given a cluster center $c_i$, we generate a number of random points according to a multivariate normal distribution centered at $c_i$ with independent components each having a variance of $4$ to form cluster $i$. Out of these 300 clusters, the points in 10 randomly selected clusters are labeled positive and the rest of the points are considered as negatives. Finally, positive points are downsampled such that only $0.5\%$ of all the points are positively labeled. Aside from the unlabeled set $U_0$  which consists of $10^5$ such data points, we also generate a validation set $E$ of size $10^4$ based on which we evaluate the performance of different algorithms.

\begin{figure}
	\centering
	\includegraphics[width=0.98\linewidth]{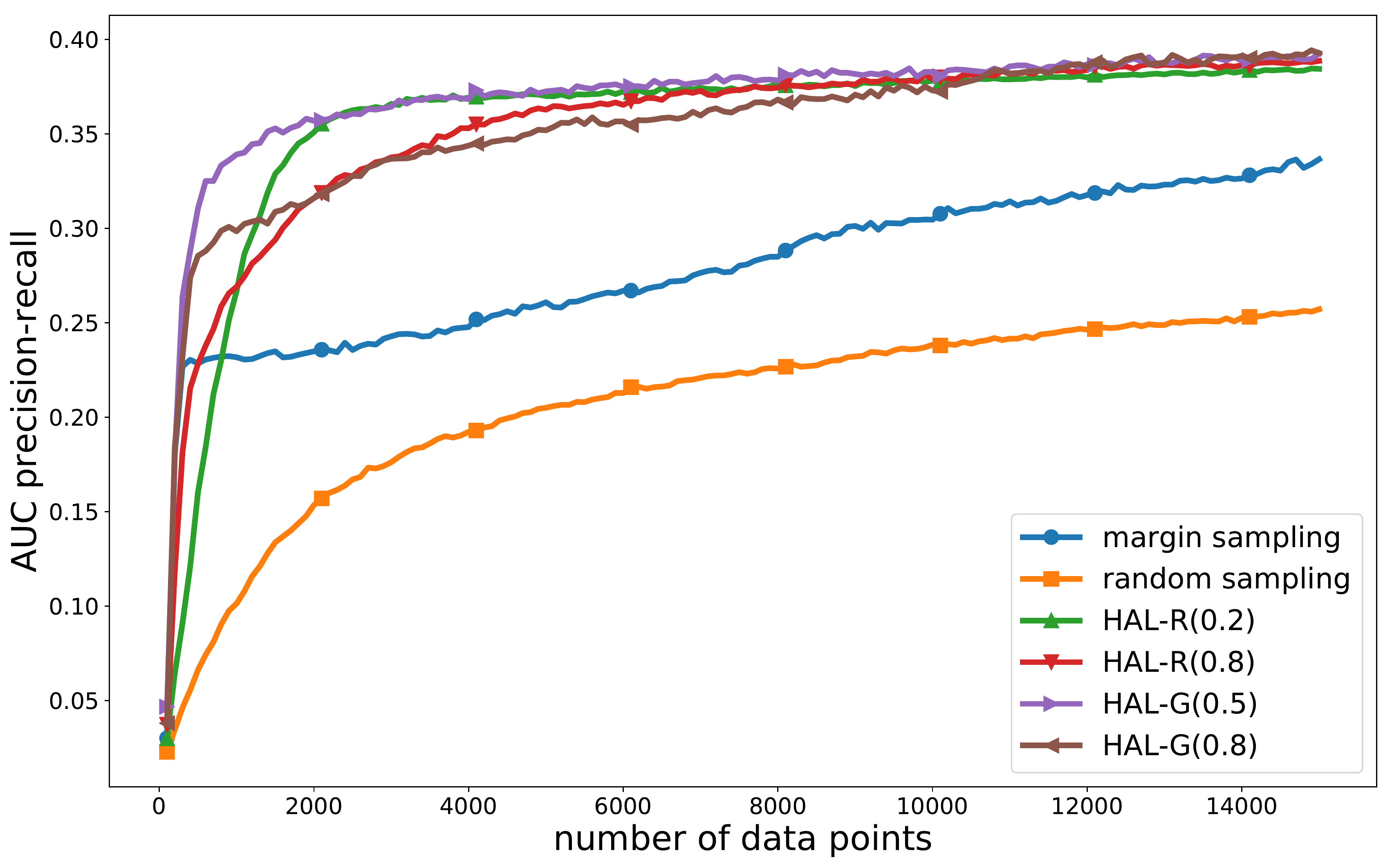}
	\caption{AUC-PR of different algorithms on a $0.5\%$-skewed synthetic data set.}
	\label{fig::synthetic_aucpr}
\end{figure}

\begin{figure}
	\centering
	\includegraphics[width=0.98\linewidth]{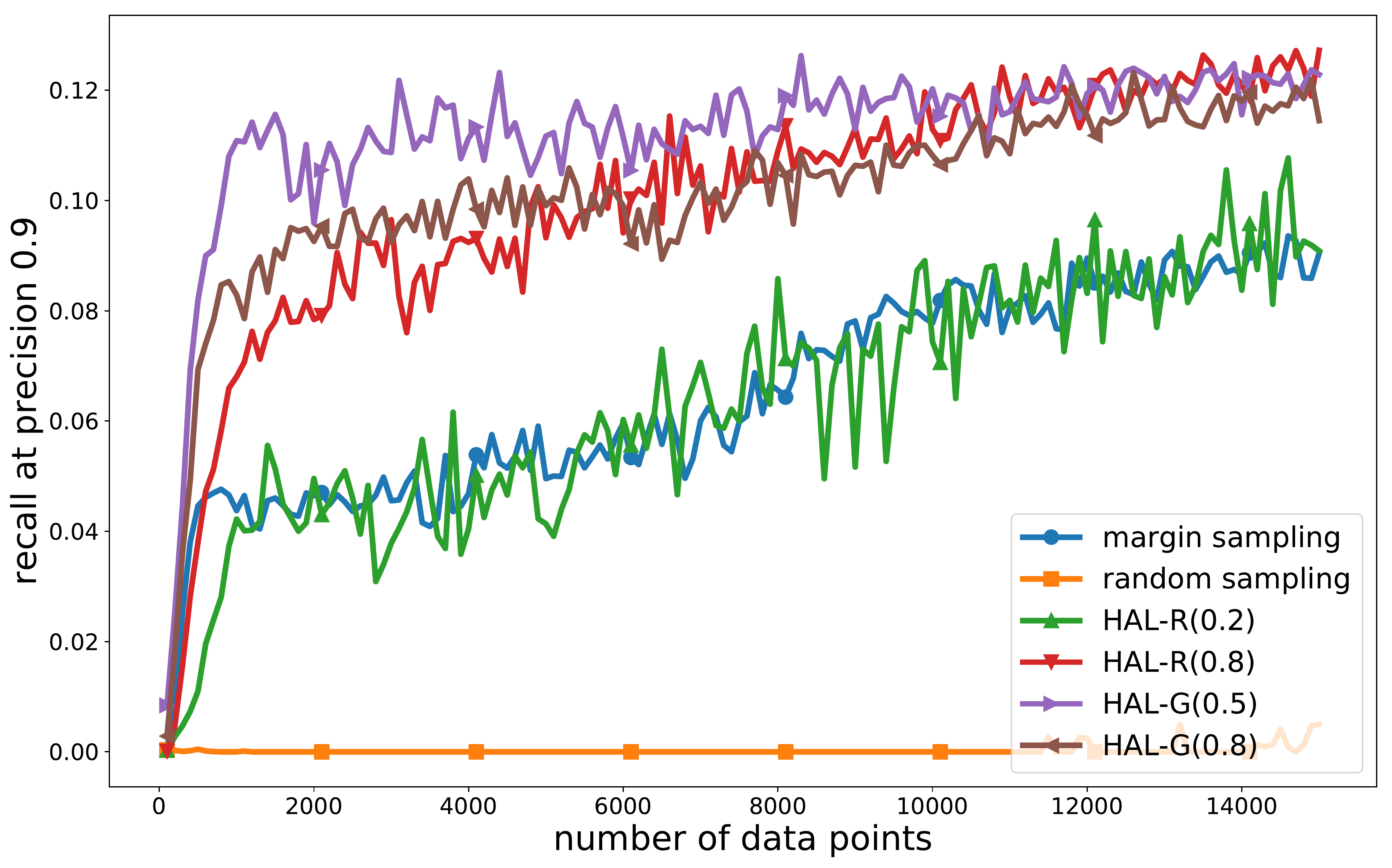}
	\caption{Recall at high precision of different algorithms on a $0.5\%$-skewed synthetic data set. }
	\label{fig::synthetic_rap9}
\end{figure}

\begin{figure}
	\centering
	\includegraphics[width=0.98\linewidth]{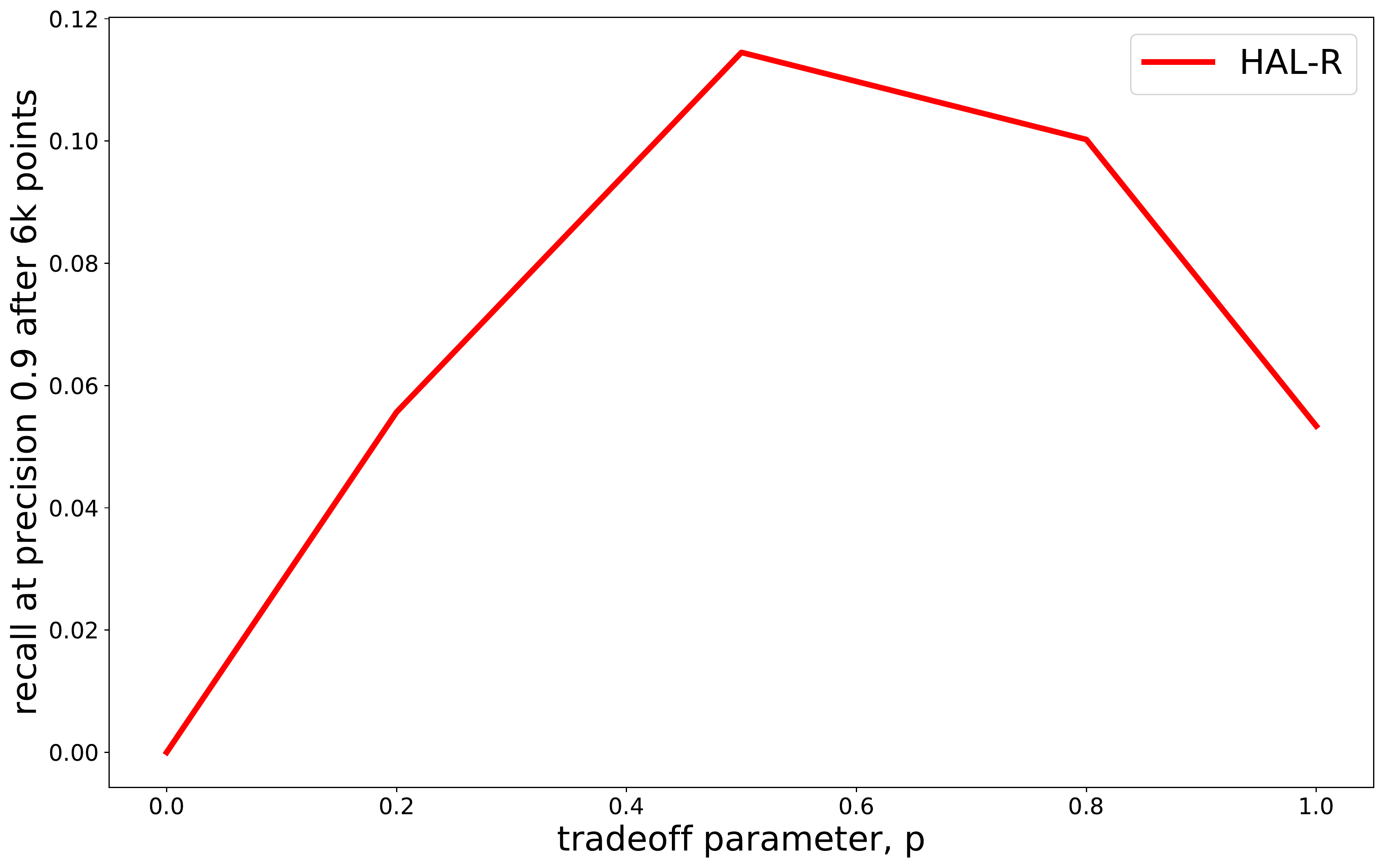}
	\caption{Recall at precision 0.9 vs.  trade-off parameter $p$ in the $0.5\%$-skewed synthetic data set. Each point is a measurement after 6k labeled data points have been observed.}
	\label{fig::synthetic_rap9_fraction}
\end{figure}

Figure \ref{fig::synthetic_aucpr} depicts AUC-PR of different algorithms versus number of labeled points. In the figure, HAL-G and HAL-R denote the hybrid algorithm with Gaussian exploration and random exploration, respectively. The number in the name refers to the fraction value, namely, the probability of using margin sampling algorithm. Each curve is obtained by averaging the results of $100$ repeated runs with the same parameters (e.g. the same exploration algorithm and trade-off parameter $p$). For each exploration scheme, we experimented with trade-off parameters $p$ from 0 to 1.0 with step 0.1 and take the best performing fraction for comparison. The results suggest that HAL is generally better than margin sampling. In particular, HAL-G(0.5) is significantly better than the other methods at the early stage where the number of labeled data points is less than 2000. HAL-G(0.5) is about $50\%$ better than margin sampling at 2000 labeled data points. Since data labeling incurs cost, it's reasonable to evaluate the performance of an active learning algorithm as the cost (e.g. dollars) required for achieving a desired objective. Assuming the cost is proportional to the number of labeled data points, HAL-G(0.5) is 7x more effective than the margin sampling, because with less than 2000 labeled data points, HAL-G(0.5) is able to achieve the same objective as margin sampling with 14000 labeled data points. Figure \ref{fig::synthetic_rap9} plots recall at precision 0.9 achieved by different algorithms. As depicted in this figure, HAL-G outperforms HAL-R. Also HAL-R($0.8$) outperformed HAL-R($0.2$), though these two algorithms have similar performance under AUC-PR metric. Both Figures \ref{fig::synthetic_aucpr} and \ref{fig::synthetic_rap9} indicate that HAL (both for random or Gaussian exploration) with $p$ chosen to balance exploration and exploitation significantly improves over HAL-R(1.0), the plain margin sampling algorithm. Figure \ref{fig::synthetic_rap9_fraction} explores this further, by plotting the recall at precision 0.9 achieved by HAL-R for different values of the trade-off parameter $p$ after observing $6$k labeled data points. As depicted in this figure, a balanced trade-off between explore and exploit components performs much better than each component separately. 

\begin{figure}
	\centering
	\includegraphics[width=0.98\linewidth]{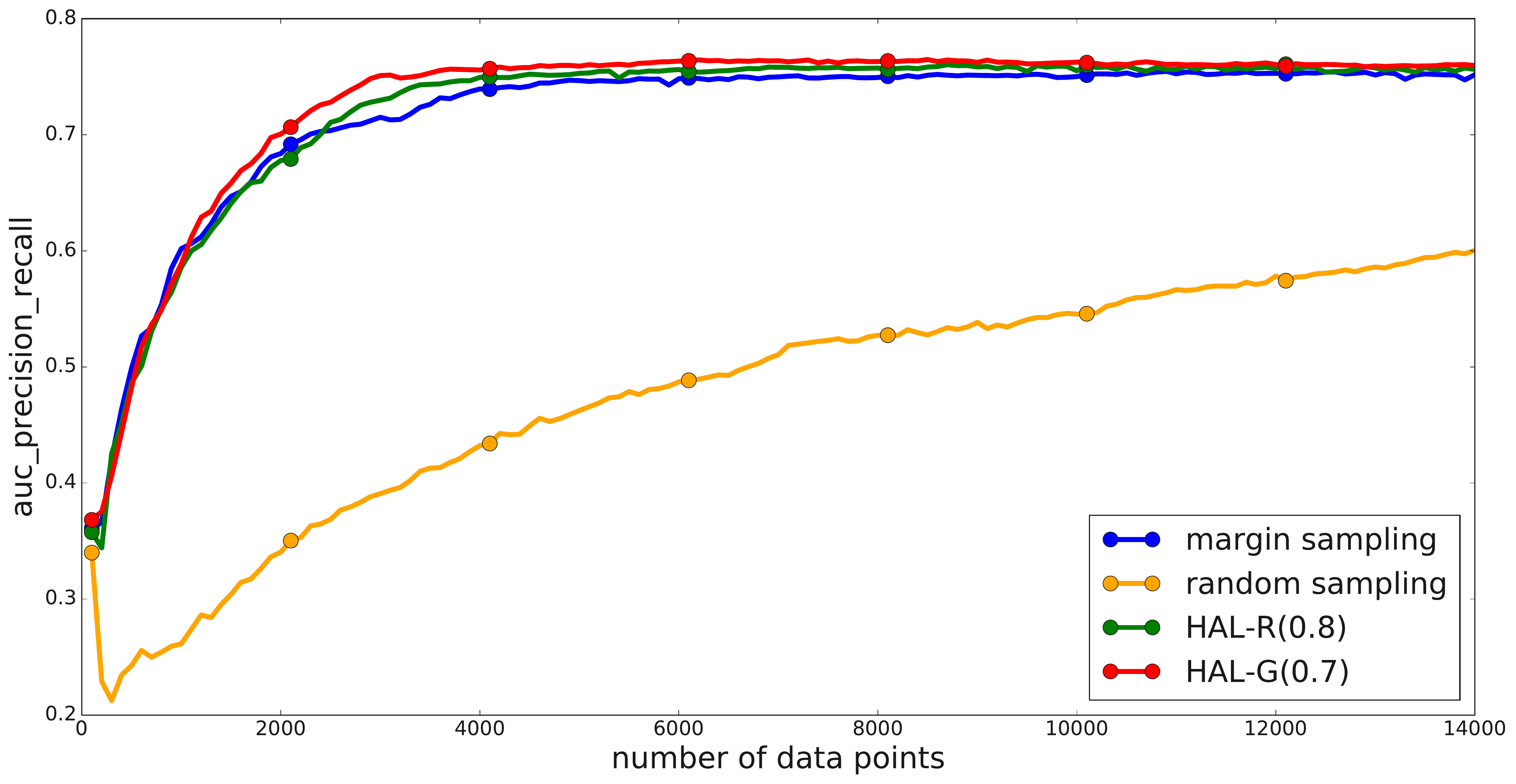}
	\caption{Area under precision-recall curve  of different algorithms on a $1.5\%$-skewed MNIST data set. }
	\label{fig::mnist_aucpr}
\end{figure}

\begin{figure}
	\centering
	\includegraphics[width=0.98\linewidth]{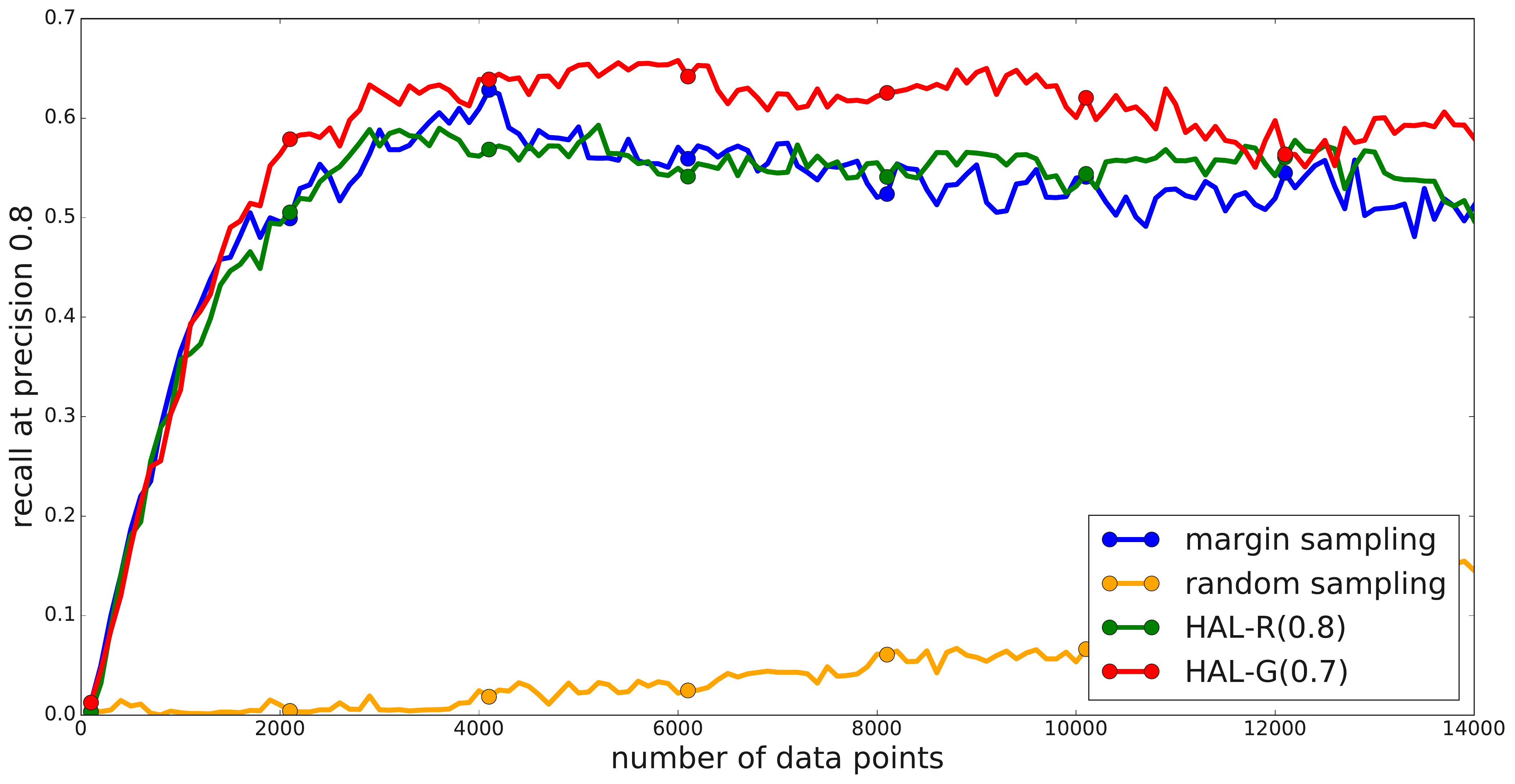}
	\caption{Recall at high precision of different algorithms on a $1.5\%$-skewed MNIST data set. }
	\label{fig::mnist_rap8}
\end{figure}

\subsection{MNIST Data Set}\label{sec::mnist_data}
To verify the above observations on real-world data sets, we performed the same experiments on the MNIST data set \cite{lecun1998gradient}. The original MNIST data set presents a fairly simple classification problem as all instances of a single class are very similar and can be distinguished from other classes via a simple clustering. To make the classification task more difficult, we turn the problem into a binary classification problem by assigning binary labels as follows:
we labeled the digits 0, 1 and 4 as positive and the other 7 digits as negatives.
Including multiple digits in the positive class ensures that there is a diversity of examples in the positive class much like with real applications in sensitive content detection etc.
Also to establish a highly skewed scenario, positive points are then downsampled to be a $1.5\%$-skewed. The initial unlabeled set and the validation set consist of $60,000$ and $10,000$ data points, respectively. 

Figures \ref{fig::mnist_aucpr} and \ref{fig::mnist_rap8} show the performance of different active learning algorithms on the modified MNIST data set. According to these results, HAL is clearly better than margin sampling at the early stage (near 3000 data points) and the advantage becomes marginal as more data points are labelled. HAL-G has a clear advantage over margin sampling in recall at high precision.
\zhaqieat{As in the previous case, we investigate the effect of the trade-off parameter $p$ on the performance of HAL in Figure \ref{fig::mnist_rap8_fraction}. This figure plots the recall at precision 0.8 achieved by HAL-R after observing $6$k labeled data points for different values of the trade-off parameter.}

\subsection{Performance with Varying Skewness}
\begin{figure}[t]
	\centering
	\includegraphics[width=0.98\linewidth]{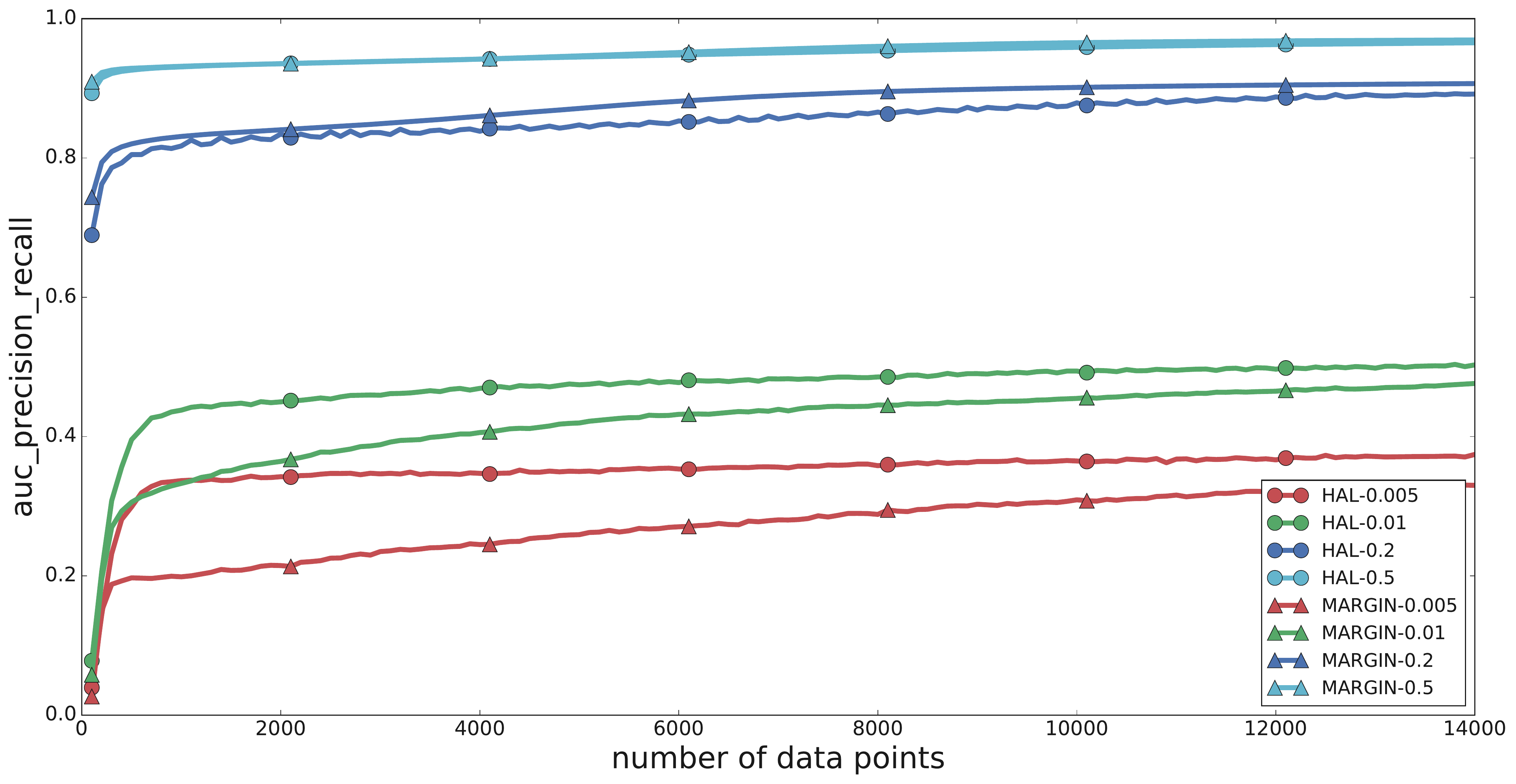}
	\includegraphics[width=0.98\linewidth]{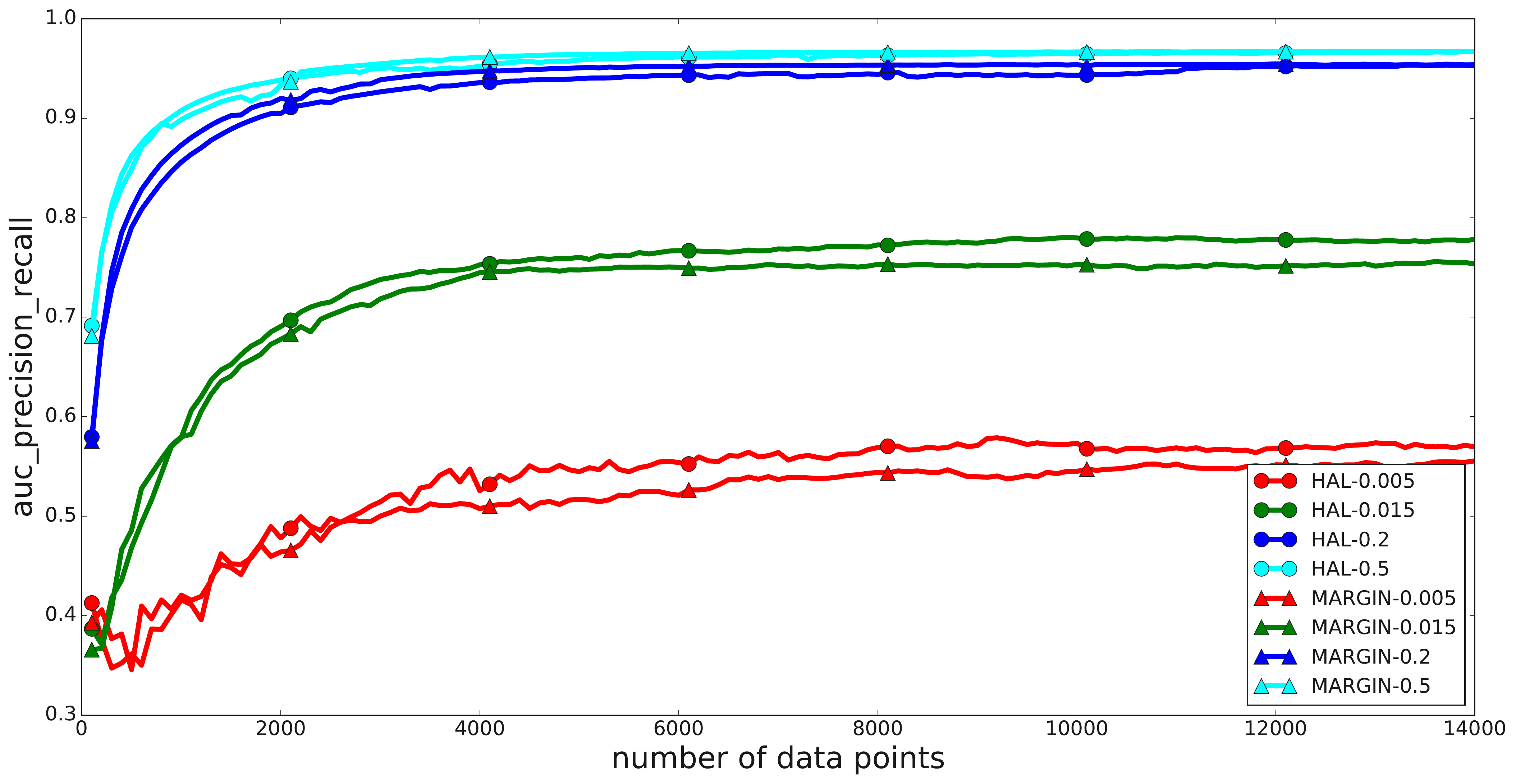}
	\caption{Area under precision-recall curve w.r.t class skewness for the synthetic (above) and the MNIST (bottom) data. HAL and MARGIN denote HAL-G and margin sampling algorithm. The number in the legend is the class skewness -- the percentage of positive labeled data.}
	\label{fig::synthetic_varying_skewness}
\end{figure}

The results on both the synthetic data set and real-world data set demonstrate that the HAL algorithm is advantageous over the margin-sampler algorithm when severe class imbalance exists. To further confirm that the advantage is more pronounced when the data is more skewed, we compared HAL to the margin-sampler on class skewness varying from $0.005$ to $0.5$ on both the synthetic and MNIST data. The setting is the same as the settings in  Section~\ref{sec::synthetic_data} and Section~\ref{sec::mnist_data} except the skewness. The results are summarized in Figure~\ref{fig::synthetic_varying_skewness}.

In the figure, lines in the same color correspond to the results of $HAL$ and margin sampling algorithms for the same skewness. The skewness is indicated as the number in the legend and the smaller the number, the more skewed the label distribution is. The gap between the same colored lines reflects the advantage of one method over the other. As seen from the results for the synthetic data, when the class skewness is severe, e.g. $0.005$ and $0.01$, HAL-G consistently outperforms the margin sampling at any given number of labeled data points. It's worth noting that HAL-G achieves an almost saturated objective value at an early stage (with less than 2000 labeled data points), which gives it a huge advantage over the margin sampling algorithm. As more data points are labeled, margin sampling algorithm starts to catch up HAL-G, but still its performance at 14000 labeled data points is slightly worse than the performance of HAL-G at less than 2000 labeled data points. In other words, HAL-G is 7x more effective than margin sampling.

At a given number of labeled data points, 2000 for example, the gap between two same colored lines gets smaller when the label skewness reduces and the gap completely disappears when the class is perfectly balanced (skewness is $0.5$). This observation shows that HAL-G is more advantageous when the label is more skewed. The results for the MNIST data shows a similar pattern.

\section{Discussion and Future Work}
\label{sec::discussion}
We discuss three aspects of deploying HAL on real-world data sets
that open up additional avenues for investigation -- \zhaqieat{choosing the right objective for model training,} designing a policy for building up a validation set and designing more principled exploration schemes.

\zhaqieat{
\subsection{Choosing the Right Objective}
In our experiments, we optimized for the cross entropy
loss, which for a binary classification task reduces to a log-loss. 
Figure~\ref{fig::mnist_rap9} shows the recall at precision of 0.9 for different variants we compared
over the MNIST data set. In contrast to the AUC-PR
metric (Figure~\ref{fig::mnist_aucpr} and recall at precision 0.8 (Figure~\ref{fig::mnist_rap8}), this metric increases initially to over 0.30, and then gets worse as we gather more labeled points.

While we are aware that optimizing for the log-loss is 
not the same as optimizing for recall at a specific precision, we were disappointed by this result.
We plan to explore techniques that allow us to directly optimize for objectives like recall at precision 0.9.
Techniques such as the ones in~\cite{eban2016scalable}
describe ways to construct approximations of non-decomposable objectives that can be optimized
using stochastic gradient descent.
The HAL framework can be applied irrespective of the objective function that the classifier is optimizing for.
We expect this is likely to correct the discrepancy between what we observed for a metric like AUC-PR, which averages the performance over the entire precision-recall curve versus metrics that are more directly relevant to our application, such as recall at a high precision threshold.
}
\subsection{The Choice of the Fraction value $p$}
Our proposed hybrid algorithm is parameterized by fraction value $p$. $p$ is a hyper-parameter which can't be learned from data. As seen from the results on the synthetic and MNIST datasets, the choice of fraction value has a significant impact on the algorithm's performance. Though we're able to identify an optimal value for $p$ by trivially examining various values of $p$ in our experiments, we have to admit that this approach might be limited for a real-world problem. In the future work, we want to explore methods which can set the fraction value dynamically.

\subsection{Absence of Validation Set}
In our experiments, the target metrics for the model at each step
are reported on a validation set.
The existence of a such a validation set is necessary for comparing
various active learning algorithms.
In practice, however, such a validation set might not exist.
Recall that in our problem setting, training data is already sparse.
Consequently, additional effort needs to be expended labeling examples to 
construct a validation set.
Then, at every step, in addition to considering which points should be selected
for labeling and inclusion in the training set, we also need to consider if
we should label points and add them into the validation set.
This is a non-trivial problem given that the data set is extremely skewed.

Choosing a good validation set is important since that may determine whether
a model is used in production.
Having a poor validation set may mean a classifier that detects sensitive
content does worse in production than expected -- this is clearly undesirable.
Choosing a cost-effective strategy for constructing such a validation set is
an open research question.
For example, if we sample randomly, we are likely to end up with the majority of selected points being negatives.
In this case, a large number of samples is needed to establish a meaningful validation set. This may prove to be very expensive.
Two questions naturally arise: 1) What is a good strategy for sampling
points to construct a validation set, and 2) Given a budget, what's the right allocation
between choosing a point for inclusion in the training set vs. the validation set.
We believe these are interesting problems for further research.

\zhaqieat{
\begin{figure}
	\centering
	\includegraphics[width=0.98\linewidth]{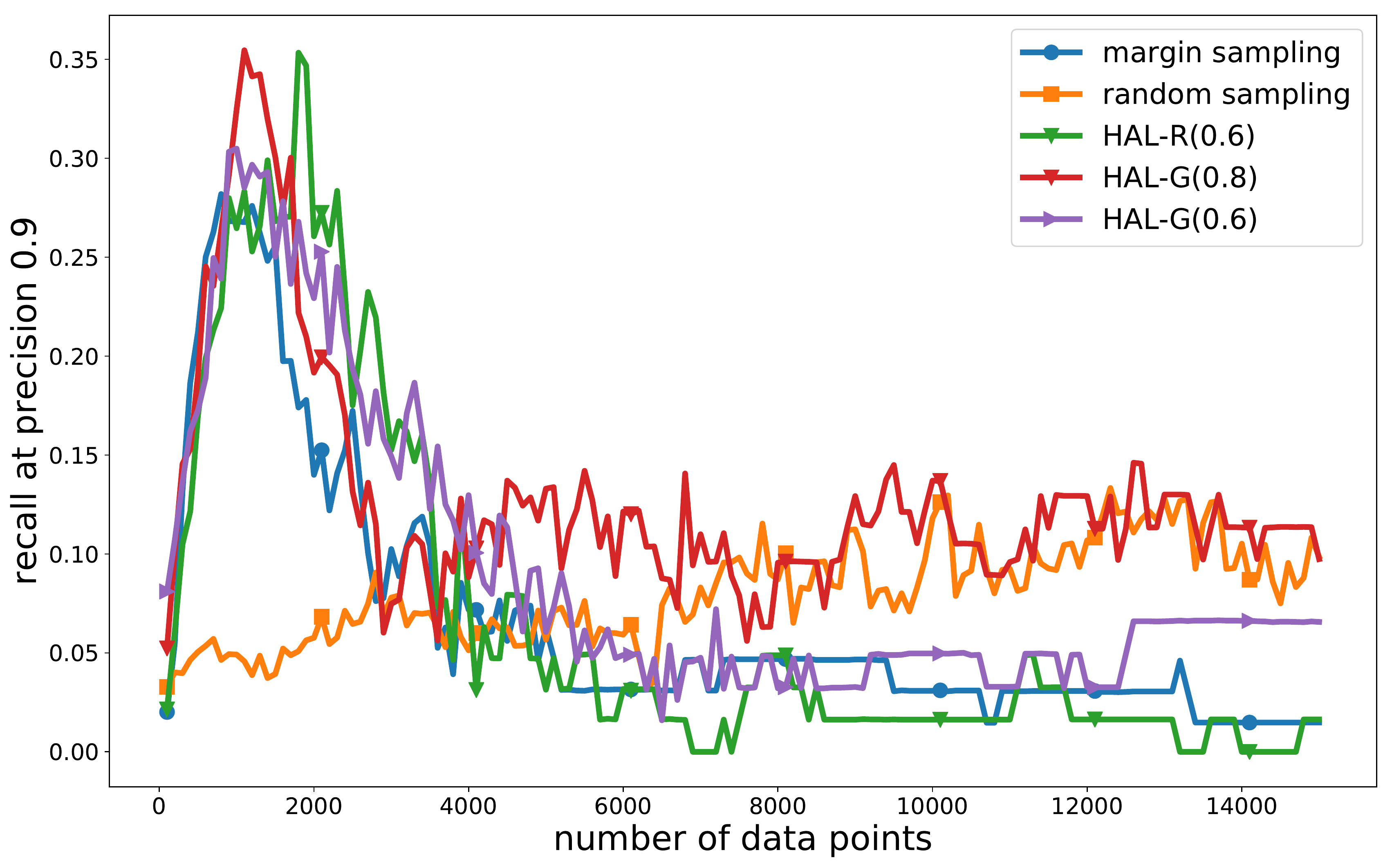}
	\caption{Recall at high precision 0.9 on MNIST data set for different $p$ values. The performance increases initially and then decreases.}
	\label{fig::mnist_rap9}
	\vspace{-5mm}
\end{figure}
}

\subsection{Exploration Schemes}
Our Gaussian exploration simply estimates the label utility for an unlabeled point by considering its distance to existing labeled points and assuming label smoothness in the feature space. Such an exploration scheme prefers selecting points far away from the labeled points and encourages diversification in the selected points when the algorithm proceeds with small batch size. The exploration is quite relevant to the sensor location selection problems in \cite{guestrin2005near, krause2007nonmyopic}. In their problem, the goal is to identify the best $k$ locations to deploy the sensors so that the spatial measurement (e.g. river pH value) can be estimated with least uncertainty. In \cite{guestrin2005near}, the authors model the spatial measurement as a Gaussian process with known parameters and select the points based on mutual information criteria that most reduce the uncertainty of the unlabeled points. In their approach, the Gaussian process allows modeling uncertainty in a natural manner. While our approach uses deep neural networks for the classification, modeling uncertainty for unlabeled points is less intuitive. In future work, it would be interesting to explore new  exploration schemes based on more principled criteria like mutual information used in \cite{guestrin2005near}.
Our exploration scheme also doesn't distinguish between
the utility of closeness to a positive labeled point or a negative labeled point.
Incorporating a preference for positive labels
is also an interesting avenue for further investigation.

\section{Conclusions}
\label{sec::conclusions}
In this paper, we considered active learning for a practical binary classification task two distinctive features: skewed classes and scarce initial trainind data. We proposed Hybrid Active Learning (HAL) as a modular solution to the active learning problem in such a setting which trades off between exploit and explore components. While we have only considered margin sampling as the exploit component of our algorithm, its explore component can be designed in many different ways. Through simulation results on synthetic and real-world data, we showed that our hybrid algorithm, even with simple exploration schemes such as random and Gaussian exploration, significantly improves over the baselines.

We identified several topics for future work: \zhaqieat{adaptively varying the
parameter which trades off between the exploit and explore components,}
choosing a good strategy for building a validation set, and
trying other exploration schemes. While this paper has focused on a binary classification task, we expect to adapt this approach to multi-class classification and regression tasks.

\bibliographystyle{ACM-Reference-Format}
\balance
\bibliography{references}

\end{document}